\documentclass{article}
\usepackage{spconf,amsmath,graphicx}

\usepackage{amssymb}
\usepackage{subfig}
\usepackage{pbox}
\usepackage{url}

\DeclareMathOperator*{\argmax}{arg\,max}

\title{Class-Specific Image Denoising using Importance Sampling}
\name{Milad Niknejad, Jos\'e M. Bioucas-Dias, M\'ario A. T. Figueiredo\thanks{The research leading to these results has received funding from the European Union's  Seventh Framework Programme (FP7-PEOPLE-2013-ITN) under grant agreement 607290 (SpaRTaN).}}
\address{Instituto de Telecomunica\c{c}\~oes, and \\
Instituto Superior T\'ecnico, Universidade de Lisboa, Portugal\\}
\begin{document}
%\ninept
%
\maketitle
\begin{abstract}
In this paper, we propose a new image denoising method, tailored to specific classes of images, assuming that a dataset of clean images of the same class is available. Similarly to the \textit{non-local means} (NLM) algorithm, the proposed method computes a weighted average of non-local patches,  which we interpret under the importance sampling framework. This viewpoint introduces flexibility regarding the adopted priors, the noise statistics, and the computation of Bayesian estimates. The importance sampling viewpoint is exploited to approximate the \textit{minimum mean squared error} (MMSE) patch estimates, using the true underlying prior on image patches. The estimates thus obtained converge to the true MMSE estimates, as the number of samples approaches infinity. Experimental results provide evidence that the  proposed denoiser outperforms the state-of-the-art in the specific classes of face and text images.
\end{abstract}
\begin{keywords}
Patch-based image denoising, class-adapted denoising, non-local means, minimum mean squared error, importance sampling. 
\end{keywords}
\section{Introduction}
\label{sec:intro}
Image denoising is one of the classical and fundamental problems in image processing and computer vision. In the past decade, the state-of-the-art has been dominated by patch-based methods, not only in image denoising, but also in more general inverse problems. In some approaches (called \textit{internal}), the image is denoised using information exclusively extracted from the noisy image. For example, denoising is carried out by averaging similar patches (as in NLM \cite{2005_buades_nonlocalmenas}), by collaboratively filtering sets of similar patches (as in BM3D \cite{2007_dabov_bm3d}), by learning a \textit{Gaussian mixture model} (GMM) from the noisy image and then using it a prior to obtain MMSE patch estimates \cite{Teodoro2015}, or by obtaining \textit{maximum a posteriori} (MAP) patch estimates using a Gaussian prior estimated from a collection a similar patches \cite{Niknejad_TIP2015}. The so-called \textit{external} methods take advantage of a dataset of clean image patches, which can be used in different ways: to denoise each noisy patch by computing weighted averages of clean patches \cite{2011_zontak_internal};  to learn a prior for clean patches (\textit{e.g.}, a GMM  \cite{2011_zoran_learning}), which is subsequently used to denoise the noisy patches \cite{2005_roth_fields}. Hybrid external/internal methods have also been proposed \cite{2013_mosseri_combining}.

Let  $\mathbf{y} ={\bf x + n} \in \mathbb{R}^{p}$, where ${\bf n}\sim{\cal N}({\bf 0},\sigma^2{\bf I})$ is additive Gaussian noise of variance $\sigma^2$, denote a noisy observed image patch and $\mathbf{x} \in \mathbb{R}^{p}$ the corresponding clean  patch. Some external image denoising methods estimate  $\mathbf{x}$  via the non-parametric weighted average 
\begin{equation}
\label{eq:selfim}
\widehat{\mathbf{x}}=\frac{\sum_{j=1}^n{w_{j}\mathbf{z}_j}}{\sum_{j=1}^n{w_{j}}},
\end{equation}
where $\{\mathbf{z}_j, j=1,..,n\}$ is a set of clean patches selected from an external
dataset, $w_{j} = \exp(-\tfrac{1}{2\sigma^2}||\mathbf{y}_i-\mathbf{z}_j||_2^2)$.
If $\{\mathbf{z}_j, j=1,..,n\}$ is a set of samples from a prior  
$p_X$, then 
\[
\lim_{n\rightarrow\infty}\widehat{\mathbf{x}} = \mathbb{E}[\bf x|y],
\] 
\textit{i.e.}, as $n\rightarrow\infty$, $\widehat{\mathbf{x}}$ converges to the MMSE estimate of $\bf x $ under that prior \cite{2011_levin_natural}. However, computing  \eqref{eq:selfim} using all the patches in some large external dataset is computationally very demanding.  In order to mitigate this computational hurdle, \textit{$k$ nearest neighbours} ($k$-NN) clustering  has been used \cite{2011_zontak_internal} to find similar patches and thus to reduce the number of patches averaged in \eqref{eq:selfim}. However, given that the clustering is performed on noisy patches, its quality is often questionable.
 
The success of patch-based denoising methods relies crucially on the suitability of the priors used. 
Although several studies provide evidence that leptokurtic multivariate distributions are a good fit to image patches \cite{2015_gerhard_modeling}, those densities have seldom been used for denoising, due to algorithmic hurdles raised by the learning procedure and posterior inference. 

In many applications, the noisy image is known to belong to a specific class, such as text, face, or fingerprints, and this knowledge should be exploited by the denoising method.  One approach to implement this idea is to use an external method, based on a dataset of clean images from the specific class in hand, rather than a general-purpose dataset of natural images  \cite{2015_luo_adaptive,2016_teodoro_image}.  The obvious rationale is that more similar patches can be found in the external set of images from the same class than in a generic dataset, and the statistical properties of the patches derived from the class-specific dataset are also better adapted to the underlying clean image.

In this paper, we first show that the non-parametric formula in (\ref{eq:selfim}) can be derived from the \textit{importance sampling} (IS) framework \cite{Hesterberg}, which is a method of the Monte-Carlo family \cite{Robert_Casela_1999}. Then, based on the IS perspective, we propose an image denoising method using class-specific external datasets, with two stages: in the first stage, a set of multivariate \textit{generalized Gaussian} (GG) distributions is learned from the external clean patches; then, noisy patches are denoised by first assigning each to one of the learned GG distributions, and then approximating the MMSE estimate via (\ref{eq:selfim}).   Under the IS framework, the MMSE patch estimates are approximated by sampling directly from the patches from which the GG distributions were estimated. The obtained results show that the proposed method outperforms other state-of-the-art general and class-specific  denoisers.

In the following sections, we first describe the IS viewpoint for (\ref{eq:selfim}). Then, the proposed method for class-specific image denoising is described. Finally, experimental comparisons with the state-of-the-art methods are conducted.

\section{Importance sampling}
A fundamental step in a patch-based denoising algorithm is the estimation of the clean patches from the noisy  ones. A classical result in Bayesian point estimation is that the MMSE estimate is given by the posterior expectation \cite{Robert1994}:
\begin{equation}
\label{eq:int}
\mathbb{E}[\mathbf{x}|\mathbf{y}]= \! \int \mathbf{x} \,p_{X|Y}(\mathbf{x}|\mathbf{y}) \; d\mathbf{x}
= \! \int  \mathbf{x} \, \frac{p_{Y|X}(\mathbf{y} | \mathbf{x})\; p_{X}(\mathbf{x})}{p_Y(\mathbf{y})}\; d\mathbf{x},
\end{equation}
where $p_{X}$ is the prior and $p_{Y|X}$ is the likelihood function, and the second equality is simply Bayes' law. Computing (\ref{eq:int}) is usually far from trivial, except when a conjugate prior is used \cite{Robert1994}; a famous example is the Gaussian (or mixture of Gaussians) prior with a Gaussian likelihood, for which the posterior expectation has a simple closed-form. 

One way to approximate \eqref{eq:int} is to simply average random samples ${\bf x}_1,\dots,{\bf x}_n \sim p_{X|Y}$. However, sampling from $p_{X|Y}$ may not be a simple task. In particular, its normalization constant $p_Y(\mathbf{y})$ is itself hard (or impossible) to compute, as it is itself an integral that is intractable for arbitrary priors.  

One way to circumvent the difficulty in sampling from  $p_{X|Y}$ is to resort to \textit{importance sampling} (IS) \cite{Hesterberg,Robert_Casela_1999}. By invoking the  law of large numbers, $\mathbb{E}[\mathbf{x}|\mathbf{y}]$ can be approximated by averaging $\mathbf{x}_i \frac{p_{Y|X}(\mathbf{y}|\mathbf{x}_i)}{p_Y(\mathbf{y})}$  using 
random samples ${\bf x}_1,\dots,{\bf x}_n\sim p_X$. Since the marginal density $p_Y(\mathbf{y})$ is still unknown, we may resort to the so-called \textit{self-normalized IS} (SNIS), which does not require knowledge of the normalization constants of target density $p_{X|Y}$ \cite{Hesterberg,mcbook}:
\begin{equation}
\label{eq:selfimp}
\widehat{\mathbb{E}}_n [ \mathbf{x}|\mathbf{y}] = \frac{\displaystyle\sum_{j=1}^n{\mathbf{x}_j\, p_{Y|X}(\mathbf{y}|\mathbf{x}_j)}}{\displaystyle \sum_{j=1}^n{p_Y(\mathbf{y}|\mathbf{x}_j)}},
\end{equation}
where $\mathbf{x}_{1},...,\mathbf{x}_n$ is a set of independent samples drawn from $p_X$. It can be shown that   $\lim_{n\rightarrow\infty}\widehat{\mathbb{E}}_n[\mathbf{x}|\mathbf{y}] = {\mathbb{E}}[\mathbf{x}|\mathbf{y}]$ \cite{Hesterberg}. 

Notice that \eqref{eq:selfim} is formally equivalent to \eqref{eq:selfimp}, as long as 
\[
p_{Y|X}({\bf y|x}) \propto \exp(-\tfrac{1}{2\sigma^2}\|{\bf x-y}\|_2^2),
\] 
\textit{i.e.}, if the noise is zero-mean Gaussian with variance $\sigma^2$, and the set  $\{\mathbf{z}_j, j=1,..,n\}$ in \eqref{eq:selfim} contains samples from the prior $p_X$. A special case of this denoiser, which was used to obtain a lower bound that denoising algorithms can achieve \cite{2011_levin_natural}, just averages the central pixel of the patch;  this corresponds to replacing $\mathbf{x}_{j}$ with $x_{j,c}$ in both the left and right hand sides of \eqref{eq:selfimp}, where $x_{j,c}$ denotes the central pixel of patch ${\bf x}_i$.

In \cite{2012_levin_patch}, it was shown that, for a fixed number of samples $n$,  the MSE of the estimator (\ref{eq:selfimp}) for the central pixel is reduced if the variance of samples, given the noisy patch, is decreased. Since the use  of patch samples from a clean dataset, with a given (but unknown) distribution, tends to have a large  variance, $n$ has to be large in order to obtain  a good approximation of $\mathbb{E}[\mathbf{x}|\mathbf{y}]$,  as reported in \cite{2011_levin_natural}. Aiming at reducing the sample size in \eqref{eq:selfimp}, some authors  use only a subset of $k$ clean patches that are the most similar to the noisy patch \cite{2011_zontak_internal}. However, because this approach compares a noisy patch with clean patches, this subset is not guaranteed to contain a proper set of similar and correlated patches.

In this paper, the large sample size hurdle is alleviated by sampling from clusters of clean patches obtained  from class-specific datasets. The obtained clusters, based on GG densities, have  low  intra-cluster variance, which is equivalent to strong correlation among the samples in a given cluster.

\section{Proposed method}

\subsection{Learning patch priors}
Learning image priors is an important step in many image denoising algorithms.  In \cite{2005_roth_fields}, a Markov random field is learned from a set of natural images whose potentials are modelled as a \textit{product of experts} (PoE). In the EPLL approach \cite{2011_zoran_learning}, a mixture of multivariate Gaussians is learned from the clean patches in an external dataset. In \cite{Teodoro2015,2012_yu_solving,2015_niknejad_gmm}, a mixture of Gaussians is learned from the patches of the noisy image. In this work, we fit a set of $M$ multivariate GG densities to a set of clean patches of the class-specific external dataset. The GG  density with parameters $\Theta=\{\boldsymbol{\mu},\mathbf{\Sigma},\beta\}$ has the form
\cite{2013_pascal_parameter}
\begin{equation}
\label{eq:GGD}
p_X(\mathbf{x};\Theta)=\frac{\beta \Gamma(\frac{p}{2})}{\pi^{\frac{p}{2}}\Gamma(\frac{p}{2\beta})2^{\frac{p}{2\beta}}|
	   \mathbf{\Sigma}|^{\frac{1}{2}}}e^{-\frac{1}{2}((\mathbf{x}-\boldsymbol{\mu})^T\mathbf{\Sigma}^{-1}(\mathbf{x}-\boldsymbol{\mu}))^\beta},
\end{equation}
where $\beta>0$ is the shape parameter, $\Gamma(.)$ represents the Gamma function, and $\boldsymbol{\mu}$ and $\mathbf{\Sigma}$ are the mean vector and the covariance matrix,
 respectively.

We take the following iterative procedure, after some initialization, and until some stopping criterion is satisfied:
\begin{enumerate}
\item cluster the clean patches using the {\em maximum likelihood} (ML) criterion  
\[
\widehat{m}_j=\argmax_{m\in\{1,...,M\}} p_X(\mathbf{x}_j|\hat{\Theta}_m),
\]
where $\hat{\Theta}_m$ is the current estimate of the parameter vector of the $m$-th cluster;

\item update the parameter estimates 
\[
\widehat{\Theta}_m = \argmax_{\Theta_m} \prod_{j: \widehat{m}_j = m} p_X( {\bf x}_j |\Theta_m),
\]
for $m=1,\dots,M $, using the method presented in  \cite{2013_pascal_parameter}.
\end{enumerate}

Notice that, although the above learning procedure  might have some computational complexity, 
it needs to be applied just  once, for a given class-specific image dataset.

\subsection{Image denoising}
In order to denoise the degraded image, each noisy patch is assigned to one of the distributions learned from the external dataset  by computing
\begin{equation}
\widehat{m}=\argmax_m p_Y(\mathbf{y}|m), \label{eq:assignpatch}
\end{equation}
where $p_Y({\bf y}|m)$ is the density of  $\mathbf{y}=\mathbf{x}+\mathbf{n}$, under the
prior $p_X({\bf x}; \widehat{\Theta}_m)$. If the cluster densities  of the clean patches were Gaussian, then
$p_Y(\cdot|m)\sim{\cal N}(\boldsymbol{\mu}_m,\mathbf{\Sigma}_m+\sigma^2{\bf I})$
\cite{2011_zoran_learning}. However, since we are using GG densities, the densities $p_Y({\bf y}|m)$ do  not have a simple expression, making it impractical to compute the ML assignments \eqref{eq:assignpatch}. At this point, we take a pragmatic  decision: we fit Gaussian densities to the   clusters, and classify each noisy patch into one of the clusters via ML using these Gaussian approximations. Compared with  GG densities, the Gaussian densities are a weaker fit; we remark, however, that they are used only for assigning the noisy patches to clusters, not in the clustering procedure itself, neither for computing the final patch estimates.

After determining the patch distribution, the MMSE patch estimates are obtained via sampling according to (\ref{eq:selfim}). 
%Generating samples from the generalized Gaussian distribution can be done by
%\begin{equation}
%\mathbf{x}=\boldsymbol{\mu}_m+\tau \mathbf{\Sigma}_m^{\frac{1}{2}} \mathbf{u},
%\end{equation}
%where 
%\begin{equation}
%\tau^2 \sim \Gamma\left(\frac{p}{2 \beta},2\right),
%\end{equation}
%with  $\Gamma(a,b)$ denoting the univariate gamma distribution with  parameters $a$ and $b$, and $\mathbf{u}$ a random vector uniformly distributed in the unit sphere 
Although it is possible to sample efficiently from a GG density \cite{2013_pascal_parameter}, we follow an alternative approach: we use as samples in (\ref{eq:selfim}) a set of randomly chosen clean patches from the cluster of clean patches to which it is assigned as explained in the previous paragraph. In this way, we are sampling from the underlying patch distribution, rather than from any fitted parametric density. As already mentioned, the reason for  clustering is to reduce the variance of the samples used in the importance sampling formula.

\subsection{More improvements}

Although the importance sampling viewpoint for the formula in (\ref{eq:selfim}) brings flexibility to the estimation of the clean patches, it also has well known shortcomings. One of them  is the large variation of the importance weights $w_j$'s, which is associated with samples of very low representativity; this shortcoming is known as degeneracy of the weights \cite{2000_doucet_sequential}. In order to alleviate it, we use the method recently proposed in \cite{2015_koblents_population}, which simply applies the hard thresholding operator on the importance weights $w_j$'s before computing the sums in (\ref{eq:selfim}). This  prevents the samples with low values of $w_j$ to contribute to the estimate $\widehat{\bf x}_i$, and, thus, reduces the variance of the weights.

Finally, the patches are returned the original position in the image and averaged in the overlapped pixels to reconstruct the whole image. In order to further improve the algorithm performance, in the denoising step, we implement two iterations of our algorithm with a boosted image as an input of the second iteration. Boosting is a known strategy, which brings back the image details missing after the first step of denoising and that has been often used in image denoising  (\textit{e.g.}, \cite{2005_osher_iterative,2013_dong_lssc}). In this strategy, denoting the image obtained in the first stage by $\mathbf{X}^{(1)}$, the boosted image is obtained by $\mathbf{Y}^{(1)}=\mathbf{X}^{(1)}+r(\mathbf{Y}-\mathbf{X}^{(1)})$, where $r<1$ is a constant. The noise level for the second iteration is computed as $\sigma_2^2=\sigma^2-\frac{1}{N^2}||\mathbf{Y}^{(1)}-\mathbf{X}^{(1)}||_F^2$, which is exactly the same formula used in \cite{2013_dong_lssc} for the same purpose.

\begin{table*}[]
\centering
\caption{Denoising results (average PSNR over 5 test images) for  the Gore face dataset \cite{2012_peng_rasl} and the text dataset.}
\label{tb:face}
\begin{tabular}{l|l|l|l|l|l|l|l|l|}
\cline{2-9}
 & \multicolumn{2}{c|}{$\sigma=20$} & \multicolumn{2}{c|}{$\sigma=30$} & \multicolumn{2}{c|}{$\sigma=40$} & \multicolumn{2}{c|}{$\sigma=50$} \\ 
 \cline{2-9} 
 & \multicolumn{1}{|c|}{face} & \multicolumn{1}{|c|}{text}  & \multicolumn{1}{|c|}{face} & \multicolumn{1}{|c|}{text}  & \multicolumn{1}{|c|}{face} & \multicolumn{1}{|c|}{text}  & \multicolumn{1}{|c|}{face} & \multicolumn{1}{|c|}{text} \\   \hline  
 \multicolumn{1}{|c|}{BM3D}& 31.88 & 28.13&29.64&24.95&27.57&22.55&26.98&20.91  \\ \hline
 \multicolumn{1}{|c|}{EPLL (generic)}&31.66&28.15&29.43&25.21&	27.74&23.15&	26.58&21.72 \\ \hline
 \multicolumn{1}{|c|}{Class adapted EPLL}&32.34&29.23&30.16&26.39&	28.49&24.40&	27.28&23.01  \\ \hline
 \multicolumn{1}{|c|}{External Non-local means}&31.81&25.93&30.08&25.27&	28.75&	23.90&27.48& 22.50\\ \hline
\multicolumn{1}{|c|}{Luo et. al. \cite{2015_luo_adaptive}} &32.98&27.52&30.89&27.44&29.24&26.29&28.01&25.02 \\ \hline
\multicolumn{1}{|c|}{Proposed method (Gaussian)}& 32.63&30.12&30.79&27.81&29.11&26.41&27.75&25.22  \\ \hline
\multicolumn{1}{|c|}{Proposed method (GG)}& \bf{33.09}&\bf{30.93}&\bf{30.99}&\bf{28.20}&\bf{29.48}&\bf{26.75}&\bf{28.08} &\bf{25.79}  \\ \hline
\end{tabular}
\end{table*}

\section{Experimental Results}
In this section, we compare the proposed method with other internal and external image denoising methods for text and face images. We take BM3D as benchmark for internal image denoising methods. For a fair comparison, EPLL is trained with image datasets from the same class, and it is here called class-adapted EPLL. We also consider the state-of-the-art denoiser proposed by Luo et. al \cite{2015_luo_adaptive}, which was specifically designed for class-specific datasets. To compare with  methods using the non-parametric formula (\ref{eq:selfim}), we consider the external non-local means denoiser \cite{2015_luo_adaptive}.

Regarding the setting of the parameters,  in the prior learning step, the initialization is obtained by the $k$-means algorithm with $20$ clusters. The parameter $\beta$ in GG was empirically set to $0.9$. The number of patch samples used in (\ref{eq:selfimp}) was set to $n=500$. The threshold for the importance weights was set to $5\times 10^{-60}$.
 
The Gore face image dataset \cite{2012_peng_rasl} was used as the face image dataset. For the text dataset, we extracted images from the different parts of a text document with different font sizes. This way, we considered both high quality (low variance) and low-quality (high variance) image datasets. In all the experiments, 5 images of the respective dataset were randomly selected for test and the  remaining ones for training. Each test image was contaminated with white additive Gaussian noise, and then  denoised using the  methods described before. Table \ref{tb:face} reports average PSNR over the five restored images.

From those results  we may extract three conclusions: a) the proposed method outperforms the competitors, although the advantage over the denoiser in \cite{2015_luo_adaptive} is small for face; b)  the clustering using  GG prior yields better denoising results than the Gaussian one;  c) considering our method in the multivariate Gaussian case, the approximate solution to the exact distribution (our method) performs better than the exact solution to the approximate fitted distribution (EPLL method).

\begin{figure}[!t]

\subfloat[]{\includegraphics{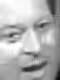}%
\label{a)}}~
\subfloat[]{\includegraphics{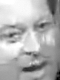}%
\label{(b)}}~
\subfloat[]{\includegraphics{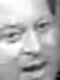}%
\label{(c)}}~
\\
\hfill
\subfloat[]{\includegraphics{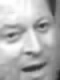}%
\label{(d)}}~
\subfloat[]{\includegraphics{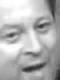}%
\label{(e)}}~
\subfloat[]{\includegraphics{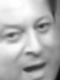}%
\label{(f)}}
\hfill
\caption{An example of Denoising for a face image in the Gore dataset ($\sigma=30$): (a) BM3D (PSNR=29.46) (b) EPLL (PSNR=28.97) (c)~Class specific EPLL (PSNR=29.91); (d) External non-local means (PSNR=31.97) (e) Luo et.~al.~(PSNR=32.20) \cite{2015_luo_adaptive}; This work (PSNR=33.02).}
\label{fig:face}
\end{figure}

\begin{figure}[!t]
\subfloat[]{\includegraphics[width=.16\textwidth]{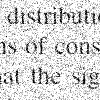}%
\label{a)}}~
\subfloat[]{\includegraphics[width=.16\textwidth]{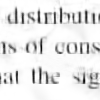}%
\label{(b)}}~
\subfloat[]{\includegraphics[width=.16\textwidth]{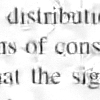}%
\label{(c)}}~
\\
\hfill
\subfloat[]{\includegraphics[width=.16\textwidth]{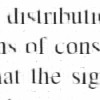}%
\label{(d)}}~
\subfloat[]{\includegraphics[width=.16\textwidth]{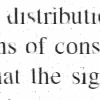}%
\label{(e)}}~
\subfloat[]{\includegraphics[width=.16\textwidth]{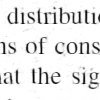}%
\label{(f)}}
\hfill
\caption{An example of Denoising for a part of text image (a) Noisy ($\sigma=60$) (b) BM3D (PSNR=20.14) (c) Class specific EPLL (PSNR=20.85); (d) External non-local means (PSNR=21.79) (e) Luo et.~al.~(PSNR=24.12) \cite{2015_luo_adaptive}; This work (PSNR=25.44).}
\label{fig:text}
\end{figure}

Two examples of denoised images in the mentioned two experiments are shown in Fig. \ref{fig:face} and Fig. \ref{fig:text}.

\section{Conclusion}
In this paper, we propose importance sampling  to approximate the MMSE estimates of clean patches in which the samples are drawn from  datasets of clean images from the same class. The clean patches were clustered under the assumption that each cluster follows a generalized Gaussian distribution. The experimental results provide  evidence that our method outperforms the state-of-the-art denoisers based on class-specific datasets. Considering other priors for image patch clustering, using the importance sampling for generic images or other noise distributions, and applying other improvements for importance sampling are subjects for future works.

\small
\bibliographystyle{IEEEbib}
\bibliography{strings,refs}

\end{document}